\documentclass[final]{cvpr}

\usepackage{times}
\usepackage{epsfig}
\usepackage{graphicx}
\usepackage{amsmath}
\usepackage{amssymb}

\usepackage[font=footnotesize,labelfont=bf]{caption}
\usepackage[belowskip=-5pt,aboveskip=3pt]{caption} % re-arrange tables on last page, select

\usepackage{multirow}
\usepackage{tabularx}
\usepackage{dsfont}
\usepackage{url}
\usepackage[tight]{subfigure}
\usepackage{color}
\usepackage{subfigure}
\usepackage{amsthm}
\usepackage[export]{adjustbox}
\usepackage{comment}

\usepackage[utf8]{inputenc} % allow utf-8 input
\usepackage[T1]{fontenc}    % use 8-bit T1 fonts
\usepackage{url}            % simple URL typesetting
\usepackage{booktabs}       % professional-quality tables
\usepackage{amsfonts}       % blackboard math symbols
\usepackage{nicefrac}       % compact symbols for 1/2, etc.
\usepackage{microtype}      % microtypography
\usepackage{enumitem}

\definecolor{dg}{rgb}{0,0.694,0.298}
\definecolor{purple}{rgb}{0.4,0.176,0.569}
\definecolor{gray}{rgb}{0.6,0.6,0.6}
\usepackage{pifont}% http://ctan.org/pkg/pifont
\definecolor{royalblue}{RGB}{65,105,225}
\usepackage[breaklinks=true,colorlinks,citecolor=royalblue,bookmarks=false]{hyperref}

\begin{document}

%%%%%%%%% TITLE
\title{Auto-Exposure Fusion for Single-Image Shadow Removal}

\author{Lan Fu\textsuperscript{1}\thanks{Lan Fu and Changqing Zhou are co-first authors and contribute equally.}, 
~Changqing Zhou\textsuperscript{2\rm$*$},
Qing Guo\textsuperscript{2}\thanks{Corresponding author: Qing Guo (\href{mailto:qing.guo@ntu.edu.sg}{tsingqguo@ieee.org})},
~Felix Juefei-Xu\textsuperscript{3},
Hongkai Yu\textsuperscript{4},\\
Wei Feng\textsuperscript{5},
Yang Liu\textsuperscript{2},
Song Wang\textsuperscript{1}\\~\\
% \author[1]{Lan Fu}
\textsuperscript{1}University of South Carolina, USA,~~ 
\textsuperscript{2}Nanyang Technological University, Singapore \\
\textsuperscript{3}Alibaba Group, USA,~~ 
\textsuperscript{4}Cleveland State University, USA,~~ 
\textsuperscript{5}Tianjin University, China

% {\tt\small lanf@email.sc.edu} 
% \and
% Qing Guo \\
% Nanyang Technological University \\
% {\tt\small qing.guo@ntu.edu.sg} \\ 
% \and
% Changqing Zhou \\
% Nanyang Technological University \\
% {\tt\small zhou0365@e.ntu.edu.sg} \\ 
% \and
% Felix Juefei-Xu \\
% Alibaba Group, USA \\
% {\tt\small  juefei.xu@gmail.com} \\
% \and
% Hongkai Yu \\
% Cleveland State University \\
% {\tt\small  hongkaiyu2012@gmail.com}
% \and
% Wei Feng \\
% College of Intelligence and Computing, Tianjin University, China \\
% {\tt\small wfeng@ieee.org}
% \and
% Yang Liu \\
% Nanyang Technology University, Singapore \\
% {\tt\small yangliu@ntu.edu.sg}
% \and
% Song Wang \\
% University of South Carolina \\
% {\tt\small songwang@cec.sc.edu}

% % \and
% % Second Author\\
% % First line of institution2 address\\
% % {\tt\small secondauthor@i2.org}
% }
}
\maketitle

%%%%%%%%% ABSTRACT
\begin{abstract}
Shadow removal is still a challenging task due to its inherent background-dependent\footnote{Background means the shadow-covered context in this paper.} and spatial-variant properties, leading to unknown and diverse shadow patterns. Even powerful deep neural networks could hardly recover traceless shadow-removed background.
This paper proposes a new solution for this task by formulating it as an exposure fusion problem to address the challenges. Intuitively, we first estimate multiple over-exposure images w.r.t. the input image to let the shadow regions in these images have the same color with shadow-free areas in the input image. Then, we fuse the original input with the over-exposure images to generate the final shadow-free counterpart. 
Nevertheless, the spatial-variant property of the shadow requires the fusion to be sufficiently `smart', that is, it should automatically select proper over-exposure pixels from different images to make the final output natural.
To address this challenge, we propose the {\bf shadow-aware FusionNet} that takes the shadow image as input to generate fusion weight maps across all the over-exposure images.
Moreover, we propose the {\bf boundary-aware RefineNet} to eliminate the remaining shadow trace further. 
We conduct extensive experiments on the ISTD, ISTD+, and SRD datasets to validate our method's effectiveness and show better performance in shadow regions and comparable performance in non-shadow regions over the state-of-the-art methods.   
We release the code in  \url{https://github.com/tsingqguo/exposure-fusion-shadow-removal}.
\end{abstract}

%%%%%%%%% BODY TEXT
%---------------------------------------------------------------------

% \vspace{-15pt}
%---------------------------------------------------------------------
%---------------------------------------------------------------------
\section{Introduction}\label{sec:intro}

% \felix{highlights: (1) multi-exposure to solve shadow problem, idea is novel, (2) overcome shortcoming, i) spatial invariant, others: shadow region color and intensity illumination degradation, position specific, color dependent. Therefore, recovering using a universal single exposure may not very well solve this problem, less flexible. ii) boundary still visible after shadow removal. Ours is much better, by using a RefineNet, with a boundary band eroded mask, to improve the boundary smoothness and consistency. 0-1 mask, shadow shifting (dark-bright) region, band-mask guided refinement. }

% image context/content covered by shadow dependent,  `shadow-covered context' --> background.

%
Shadows are present in most natural images where the light source is blocked. Spatial-variant color and illumination distortion presented in the shadow region can hinder the performance of other computer vision tasks \cite{cucchiara2003detecting, jung2009efficient, nadimi2004physical, sanin2010improved, zhang2018improving}, such as object detection and tracking, object recognition, semantic segmentation, \etc. 
%Many researches have been proposed for shadow removal to recover the underlying content.  
%

%------------------------------------
% put fig1 and part vis result for our solution
%
\begin{figure}[t]
\centering
\includegraphics[width=1.0\columnwidth]{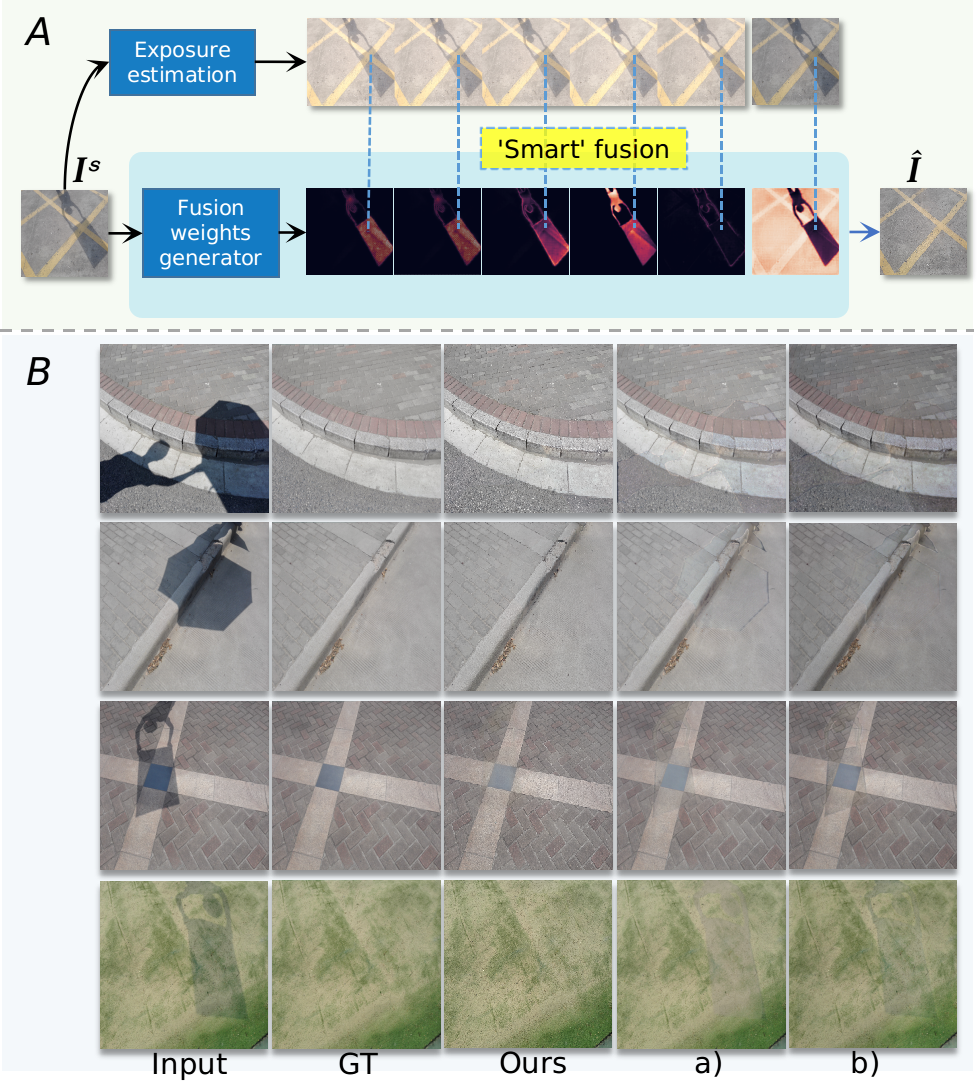}
\caption{A: Illustration of the proposed auto-exposure fusion for shadow removal. B: Visualization results of our shadow removal results with the state-of-the-art methods. a) and b) are the shadow removal results of SP+M-Net \cite{le2019shadow} and DSC \cite{hu2019direction}, respectively.}
\label{fig:fig1_vis}
\vspace{-12pt}
\end{figure}
%------------------------------------

%
Previous shadow removal works either model this task based on physical shadow models for paired shadow and shadow-free images \cite{le2019shadow} or model it as an image-to-image translation problem based on the generative adversarial networks (GAN) for unpaired shadow and shadow-free images \cite{hu2019mask}. However, the learned shadow removal transformations by GAN-based methods, \eg, MaskShadowGAN \cite{hu2019mask}, tend to generate artifacts and image blur. They also suffer from data distribution requirements, where they expect the unpaired shadow and shadow-free image sets to share statistical similarity \cite{li2019asymmetric}, which is hard to be satisfied when data acquisition is unstable. 
%此处原话是
%This
% requirement can be hard to satisfy when capturing shadow-free images is tricky,
% such as shadow-free images of urban areas [4] or moving objects [18,36].
On the other hand, the publicly available large-scale datasets of paired shadow and shadow-free images, such as SRD \cite{qu2017deshadownet}, ISTD \cite{wang2018stacked}, and ISTD+ \cite{le2019shadow}, allow shadow removal tasks to learn a physically plausible transformation in a supervised way. In this paper, we focus on paired training data to perform the shadow removal task.

Shadow casting decreases the image quality with color and illumination degradation, over-exposure of the shadow image is an effective way to enhance the image quality. Intuitively, fusing the over-exposed one and the original shadow image could obtain the desired shadow-free image. Recent shadow decomposition works \cite{le2019shadow, le2020shadow}, based on physical shadow models, mainly learn to relight the shadow image to a lit version and then fuse them together to acquire the desired shadow-free image via a shadow matte.
%improve the image quality of under/over exposed image. 
However, since shadow casting degrades the color and illumination across the spatial region in a background-dependent and spatial-variant manner (\ie, the contiguous shadow cast on the background image may cause the shadow region to appear differently based on how the original shadow-free background region looks like, as well as where the shadow is cast spatially on the background image), we argue that multiple over-exposure fusion allows much higher level of flexibility and can provide a better solution to compensate the shadow region to have the same color and illumination with its non-shadow area, and better recovers the underlying content of the shadow region.  
Shadow removal is still a challenging task for powerful state-of-the-art deep neural networks (DNN).
Unknown and diverse shadow patterns pose two challenges to existing DNN based solutions: \ding{182} Shadow removal is a background-dependent task, which requires DNN to not only recover the illumination and color consistency with the shadow free area but also to preserve the content underlying the shadow. The spatial-variant property of shadow area requires that the fusion should be `smart' enough to adaptively select the desired over-exposure pixels from various images to obtain the final shadow-free version. \ding{183} It is hard to obtain traceless background due to inconsistent shadow patterns along the boundary and inside the shadow region. 

In this paper, we propose a novel method, named auto-exposure fusion network, for single image shadow removal, as shown in Fig.~\ref{fig:fig1_vis}(A). We first utilize exposure estimation to learn multiple over-exposure images by compensating the shadow region with different exposure levels. Then we propose the \textit{shadow-aware FusionNet} in Sec.~\ref{subsec:fusion} to produce fusion weight maps across all the over-exposure images for addressing the first challenge. It can `smartly' select which over-exposed pixel is the best one to recover the position-specific background. The proposed method fuses the input image and its over-exposure versions in a pixel-wise way. Further, we propose a \textit{boundary-aware RefineNet} in Sec.~\ref{subsec:refine}, to remove the remaining shadow trace for refining the removal result obtained in the previous step. Figure~\ref{fig:fig1_vis}(B) shows that the proposed method can obtain traceless background image than the state-of-the-art methods SP+M-Net \cite{le2019shadow} and DSC \cite{hu2019direction}.
%
% from hongkai
The contributions of this paper are:
\begin{itemize}[noitemsep, nolistsep,leftmargin=*] 
    \item To the best of our knowledge, this paper is the first work to study the shadow removal problem from the perspective of auto-exposure fusion.
    
    \item To accurately remove the shadow, we propose a new learning-based shadow-aware FusionNet followed by a boundary-aware RefineNet to accurately estimate, smartly fuse, and meticulously refine multiple over-exposure maps.
    
    \item The comprehensive experimental results on the public ISTD, ISTD+, and SRD datasets show that the proposed method achieved better performance in shadow regions and comparable performance in non-shadow regions over the state-of-the-art methods.   
\end{itemize}
%

%-------------------------------------------------------------------------
% related work
%-------------------------------------------------------------------------
\section{Related Work}\label{sec:related_work}
 \textbf{Shadow removal.} Traditional shadow removal methods employ prior information, \eg, gradient \cite{gryka2015learning}, illumination \cite{zhang2015shadow, shor2008shadow, xiao2013fast}, and region \cite{guo2012paired, vicente2017leave}, for removing shadows. Recent deep learning based shadow removal methods boost the removal performance because of the available large-scale datasets of paired and unpaired shadow and shadow free images \cite{le2019shadow, ding2019argan, hu2019mask}. The Deshadow-Net \cite{qu2017deshadownet} extracted multi-context features, involving global localization, appearance, and semantics, to predict a shadow matte layer for removing shadow in an end-to-end manner. Wang \etal proposed ST-CGAN \cite{wang2018stacked} for joint shadow detection and removal by employing a stacked conditional GAN framework. The DSC \cite{zhu2018bidirectional} additionally utilized direction-aware context to improve shadow detection and removal. Le \etal \cite{le2019shadow} proposed to remove shadows from the perspective of shadow decomposition. On the other hand, the GAN based methods, \eg, MaskShadowGAN \cite{hu2019mask}, made it possible to perform shadow removal on unpaired shadow and shadow free images by viewing it as an image-to-image translation problem. However, these methods suffered from artifacts and image blur. They also required the unpaired shadow and shadow free image sets to have similar statistical distribution.
 
We model the shadow removal problem from a novel direction, \ie, an auto-exposure fusion problem on paired shadow and shadow free images. Multiple over-exposure shadow images are generated to compensate the color and illumination degradation in the shadow region, then they are `smartly' fused together to obtain the shadow free image.
%
 
%
% put framework pipline
%
\begin{figure*}[t]
\centering
\includegraphics[width=2.0\columnwidth]{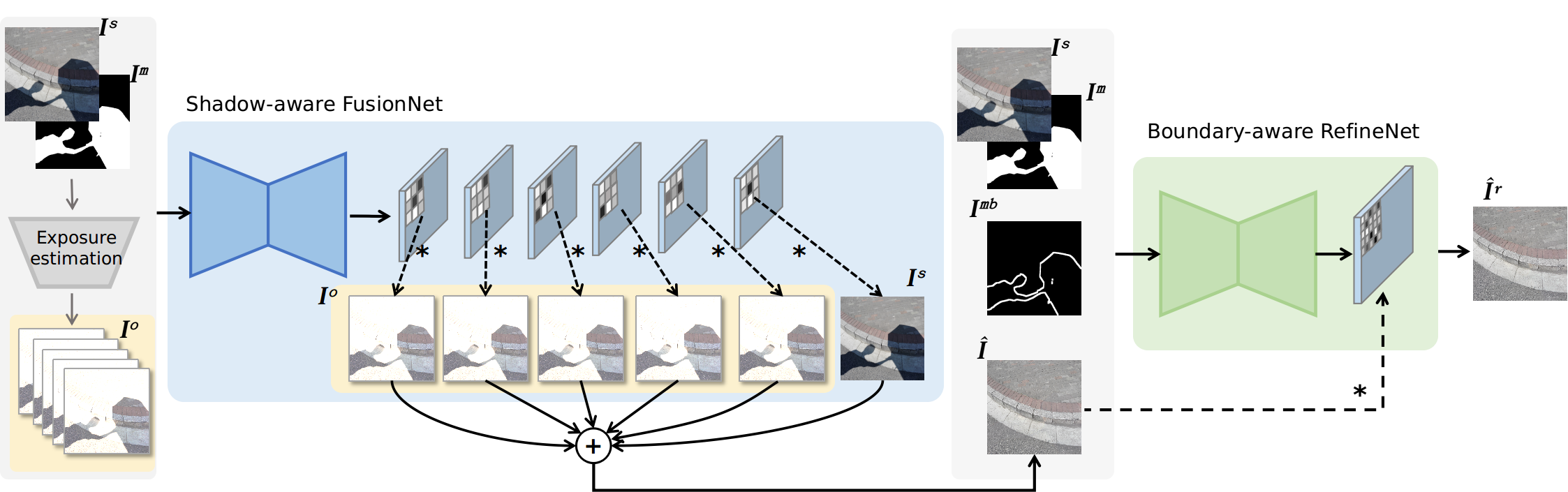}
\caption{Illustration of the proposed framework for shadow removal with shadow-aware FusionNet and boundary-aware RefineNet.}
\label{fig:framework}\vspace{-10pt}
\end{figure*}

\textbf{Exposure fusion.}
Common imaging sensors' capture range is generally limited, a picture will often turn out to be under/over exposed in real world scene. Multi-exposure image fusion (MEF) can help to refine the image quality by fusing multi-exposure images into one. MEF algorithms aim to compute the fusion weight map for each image and fuse the input sequence with a weighted sum operation. Traditional MEF methods \cite{goshtasby2005fusion, mertens2009exposure} generally performed fusion locally or pixel wisely with hand-crafted features. 
% Burt \etal \cite{burt1993enhanced} proposed a Laplacian pyramid decomposition of the image to refine the weight map for following image fusion. 
Goshtasby \etal \cite{goshtasby2005fusion} proposed a patch-wise MEF by fusing uniform blocks with the best-exposed information chosen from each image. Mertens \etal \cite{mertens2009exposure} utilized perceptual factors, such as contrast and saturation, to construct an efficient pixel-wise MEF. Li \etal \cite{li2013image} proposed a guided filter based fusion approach, taking advantage of spatial consistency, with a two-scale decomposition. Ma \etal \cite{ma2017multi} performed image fusion by optimizing a structural similarity index (MEF-SSIM) with a novel gradient descent-based method. Recent deep learning based techniques have improved the fusion performance due to high representation abilities. DeepFuse network \cite{prabhakar2017deepfuse} performed multi-exposure fusion in an unsupervised manner by employing a loss function without reference image quality. MEF-Net \cite{ma2019deep}, proposed by Ma \etal, optimized the perceptually calibrated MEF-SSIM to predict and refine the fusion weight maps. In addition to these standard MEF methods for image enhancement, recent works also discussed the effects of MEF to the image classification \cite{gao2020making,cheng2020adversarial} from the angle of adversarial attack \cite{eccv20_spark} by estimating the adversarial fusion weights with kernel prediction \cite{guo2020watch,guo2020efficientderain}.
% where the kernel prediction network \cite{guo2020watch,guo2020efficientderain} is employed to estimate 
% , which are similar with the deep network used in this work but having totally different motivation and objectives. 

In this paper, we utilize exposure fusion for the shadow removing task. Over-exposure is an effective way to enhance the image quality of shadow area. We employ pixel-wise fusion for a sequence of over-exposure images and shadow image to obtain the desired shadow-free image.
%

%-------------------------------------------------------------------------
% methodology
%-------------------------------------------------------------------------

\section{Methodology}
In this section, we propose to formulate the shadow removal as an exposure fusion problem to recover traceless background in the shadow image. We introduce the whole framework in Sec.~\ref{subsec:method_overview} and reveal the challenges. In Sec.~\ref{subsec:over_exposure}, we explain how we generate the multi-exposure images for fusion. Then, our two main contributions, \ie, \textit{shadow-aware FusionNet} in Sec.~\ref{subsec:fusion} and \textit{boundary-aware RefineNet} in Sec.~\ref{subsec:refine}, help to address the challenges and achieve much better deshadowed images.

\subsection{Exposure Fusion for Shadow Removal}\label{subsec:method_overview}
%---------Guo outline
% intuition behind exposure fusion and shadow removal.

% ref:
% The Shadow Meets the Mask:
% Pyramid-Based Shadow Removal

% challenge-> estimate multiple exposure images
% spatial-variant.
%

%---lan outline
% 1)本质 shadow region， 自己定义问题 formulation， shadow region和非shadow region区域关系
% 2)relationship to Exposure fusion， 
% 3)方案，challenge两个, smart select image to final (3.2, 3.3), 为什么做fusion， why not get the desired directly, but also has traceless, so boundary aware refinenet
% 多次曝光 处理shadow variant
% 首
%

%
We recast the shadow removal task as an exposure fusion problem and it can be formulated as
\begin{align}\label{eq:shadow}
\hat{\mathbf{I}} &=\phi(\mathbf{I}^\text{s}), 
\end{align}
where $\phi(\cdot)$ denotes a transfer function that can map the shadow image $\mathbf{I}^\text{s}$ to the corresponding shadow free image $\hat{\mathbf{I}}$. A well-exposure image, \ie, shadow free image, could be obtained by exposure fusion of brackets of multi-exposure images to improve the image quality of shadow image. The purpose of employing image over exposure is to compensate the shadow region to have the same color and illumination as the non-shadow region. In this paper, we formulate the shadow region as an under exposed area of the shadow image. Then the problem left is to recover this area to its counterpart version which has the consistent color and illumination with the unshadowed area.
Then, we can reformulate Eq.~\eqref{eq:shadow} to
\begin{align}\label{eq:expo}
\hat{\mathbf{I}} =\phi(\mathbf{I}^\text{s}, \mathbf{I}^\text{o}_{i} ), i \in \{1, 2,\ldots, N\}, 
\end{align}
where $\mathbf{I}^\text{o}_{i}$ corresponds to the $i$-th over-exposure image of shadow image $\mathbf{I}^\text{s}$. An intuitive way to solve it is to estimate an over-exposure version of the shadow image and then fuse them together to directly infer the desired shadow-free one. Nevertheless, shadow region is background-dependent and presents spatial-variant property, \ie, the color and illumination distortion across shadow region is variant, single over-exposure could not adaptively reflect the degradation in spatial space.
% give a case
%
%
% put kpn for fusionNet
%
\begin{figure*}[t]
\centering
\includegraphics[width=0.9\textwidth]{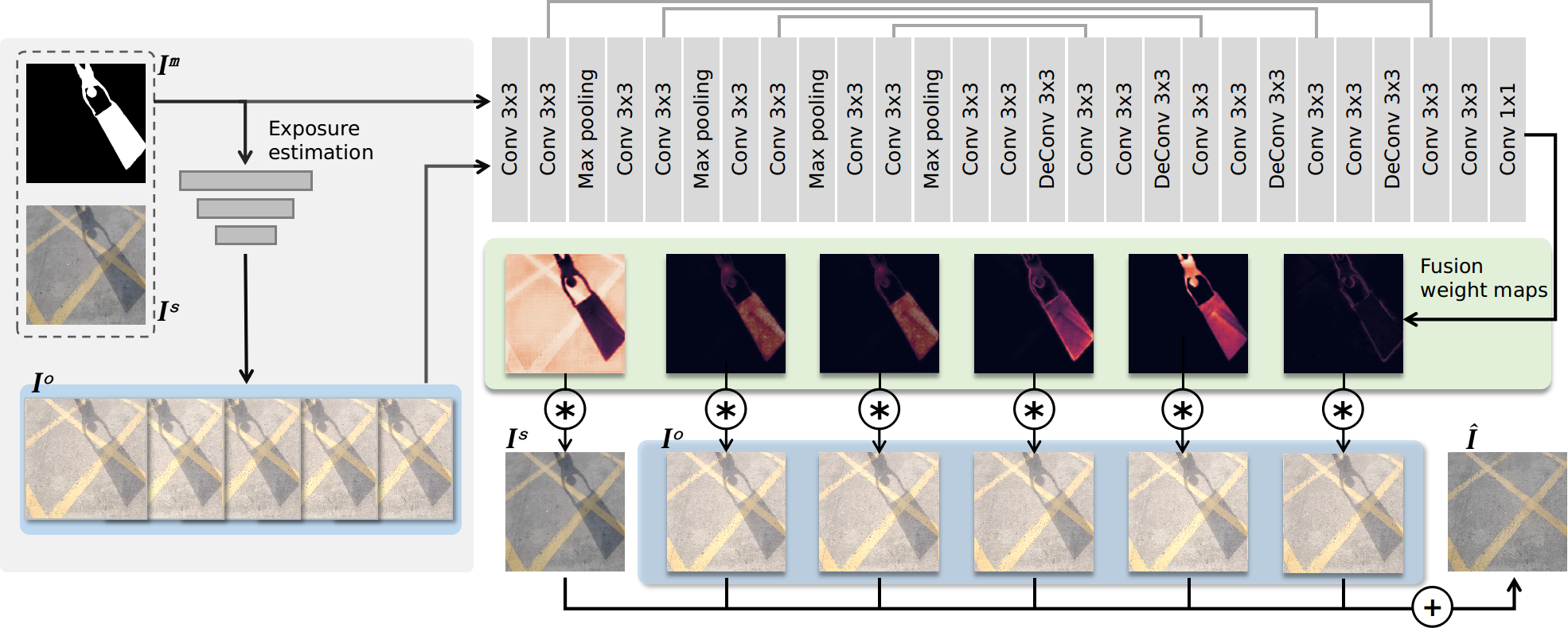}
\caption{Illustration of the proposed shadow-aware FusionNet.}
\label{fig:fusion}\vspace{-10pt}
\end{figure*}

%
%multi exposure visualization
% 
% \begin{figure}[t]
% \centering
% \includegraphics[width=1.0\columnwidth]{image/multi_expos.png}
% \caption{Illustration of the proposed multiple over-exposure estimation. The first column is the prediction, the second column is the ground-truth shadow free image and the 3-5 columns are the multiple over-exposure estimations.}
% \label{fig:multi_expos}
% \end{figure}

%
Therefore, we propose an auto-exposure fusion network for fusing shadow image with sequence of over exposed shadow images aiming to obtain the shadow free one. The whole framework of shadow removal is shown in Fig.~\ref{fig:framework}. In Sec.~\ref{subsec:over_exposure}, we employ a deep learning network to generate a sequence of over exposed shadow images. Then we propose the \textit{shadow-aware FusionNet} in Sec.~\ref{subsec:fusion} to `smartly' fuse brackets of exposed images by generating fusion weight maps across each pixel of the input image to adaptively recover the color and illumination. However, due to the existing partial shadowed region, it is hard to obtain traceless background due to the inconsistent shadow patterns along the boundary and inside the shadow area. Further, we propose a \textit{boundary-aware RefineNet} in Sec.~\ref{subsec:refine}, to remove the residual shadow trace with the help of boundary mask.

\subsection{Over-exposure Sequence Generation}\label{subsec:over_exposure}
%
% We generate multiple exposure images by multiplying the shadow image $\mathbf{I}^\text{s}$ with different scalars through the channel-wise way
We generate multiple exposure images through channel-wise weighting of the shadow image $\mathbf{I}^\text{s}$ as following:
%
%--------------------------------
\begin{align}\label{eq:exp_gen}
\mathbf{I}^\text{o}_i =\alpha_i\mathbf{I}^\text{s}+\beta_i, i \in \{1, 2,\ldots, N\},
\end{align}
%--------------------------------
%
where $\alpha_i\in\mathds{R}^{3\times 1}$ controls the exposure degree of the $i$-th over-exposure image and $\beta_i\in\mathds{R}^{3\times 1}$ decides the potential intensity shifting. To realize the goal of shadow removal, we should estimate $\{\alpha_i\}$ and  $\{\beta_i\}$ to make the shadow regions in the generated over-exposure images have the same color with the shadow-free regions in $\mathbf{I}^\text{s}$. To this end, we aim to train a DNN to estimate the exposure parameters adaptively by taking the shadow image and shadow mask $\mathbf{I}^\text{m}$ as input. Nevertheless, estimating all of the $N$ exposure parameters directly via a DNN could let the training difficult. Instead, we adopt a two-stage way: first, we train a DNN to estimate the median exposure degree, \ie, $\alpha_{\frac{N}{2}}$ and $\beta_{\frac{N}{2}}$
%
%--------------------------------
\begin{align}\label{eq:med_gen}
(\alpha_{\frac{N}{2}},\beta_{\frac{N}{2}})=\varphi(\mathbf{I}^\text{s}, \mathbf{I}^\text{m}),
\end{align}
%--------------------------------
%
where $\varphi(\cdot)$ denotes the DNN for exposure parameter estimation.
Second, we generate all exposure images by performing a simple interpolation on $\alpha_{\frac{N}{2}}$ and $\beta_{\frac{N}{2}}$ with the assumption that the over-exposure sequence's images have similar color with minor difference
%
%--------------------------------
\begin{align}\label{eq:inter_gen}
[\alpha_{i},\beta_{i}]=\gamma_i[\alpha_{\frac{N}{2}},\beta_{\frac{N}{2}}], i\in{1,2,\ldots,N},
\end{align}
%--------------------------------
%
where $\{\gamma_i\}$ denotes the interpolation coefficients. Then, the key problem becomes how to train $\varphi(\cdot)$, which is a deep regression problem. The input data of exposure estimation is the shadow image and corresponding shadow mask. The ground truth of $\alpha_{\frac{N}{2}}$ $\beta_{\frac{N}{2}}$ is calculated by performing the least squares method \cite{chatterjee1986influential} on the shadow mask covered regions of shadow image and its shadow-free counterpart. We optimize the exposure estimation by minimizing the mean squared error (MSE) between the estimated parameters and its ground truth. Note that, exposure parameters are estimated independently between color channels to adaptively adjust color distortion caused by shadow as well as camera sensor. We provide more details in the Sec.~\ref{subsec:impl}.

% we have the labeled shadow masks for training. During the inference stage,

\subsection{Shadow-aware FusionNet}\label{subsec:fusion}

% **NOTES**:
% 1. $w, b$ is channel-wise
% 2. input of relit: input image, shadow mask
% 3. U-Net256 generate kernel
% 4. refine-boundary : input includes input image, output, shadow mask, shadow mask dilate - erode, 4 parts. For now, dilate and erode use kernel 7x7 ones matrix

%
In this section, we design the FusionNet to fuse the generated over-exposure images $\{\mathbf{I}_i^\text{o}\}$ and produce the shadow-free image $\hat{\mathbf{I}}$. Intuitively, we can fuse  $\{\mathbf{I}_i^\text{o}\}$ by assigning each pixel a weight across different exposure degree
%
%--------------------------------
\begin{align}\label{eq:fusion}
\hat{\mathbf{I}}[p] = \sum_{i=0}^{N}\mathbf{W}_i[p]\mathbf{I}^\text{o}_i[p], 
\end{align}
% ~\text{subject~to}~\sum_{i=0}^{N}\mathbf{W}_i[p]=1,
%--------------------------------
%
where $\mathbf{I}_0^\text{o}=\mathbf{I}^{s}$, and $\mathbf{W}_i$ has the same size with $\mathbf{I}^\text{o}_i$. Actually, such process means that each pixel of the final shadow-free image is the linear combination of $N$ over-exposure images at the same pixel position and is fused independently. However, the fusion strategy ignores the local smoothness, leading to less natural or even noisy fusion results. Then, we further extend Eq.~\eqref{eq:fusion} by
%
%--------------------------------
\begin{align}\label{eq:fusionv2}
\hat{\mathbf{I}}[p]= \sum_{i=0}^{N}(\mathbf{K}_i\circledast\mathbf{I}^\text{o}_i)[p]=\sum_{i=0}^{N}\sum_{q\in\mathcal{N}(p)}\mathbf{k}_i^{p}[p-q]\mathbf{I}^\text{o}_i[q],
\end{align}
% &~\text{subject~to}~\sum_{i=0}^{N}\sum_{q\in\mathcal{N}(p)}\mathbf{k}_i^{p}[p-q]=1, \nonumber
%--------------------------------
%
where $\circledast$ denotes the pixel-wise convolution, \ie, each pixel is filtered by a kernel that is not shared by other pixels.
Specifically, the $p$-th pixel of $\mathbf{I}^\text{o}_i$ (\eg, $\mathbf{I}^\text{o}_i[p]$) and its neighboring pixels (\ie, $\{\mathbf{I}^\text{o}_i[q]|q\in\mathcal{N}(p)\}$) are linearly combined by an exclusive kernel (\ie, $\mathbf{k}_i^{p}$ the $p$-th kernel in $\mathbf{K}_i$) as the combination weights and $\mathbf{k}_i^{p}[p-q]$ denotes $[p-q]$-th elements of $\mathbf{k}_i^{p}$.
$\mathcal{N}(p)$ is the neighboring pixels of $p$. Compared with Eq.~\eqref{eq:fusion}, Eq.~\eqref{eq:fusionv2} considers the neighboring pixels' color and could avoid potential noisy results with better removal effect. We denote $\mathcal{K}=\{\mathbf{K}_i\}$ as pixel-wise fusion kernels.

Then, the key of generating the true shadow-free image is to estimate the fusion kernels accurately. 
Motivated by above process, we propose to estimate the fusion weight maps by training a CNN that takes the shadow image with shadow mask for guidance
%
%--------------------------------
\begin{align}\label{eq:fusionnet}
\mathcal{K}= \text{FusionNet}(\mathbf{I}^\text{m}, \mathbf{I}^\text{o}), 
\end{align}
%--------------------------------
%
where $\mathbf{I}^\text{m}$ is the shadow mask. The FusionNet is required to understand the shadow images and predict kernels that can spatially adapt to different shadow-covered contexts, thus can select suitable pixels from the multiple over-exposure images for shadow removal.

The pipeline of the shadow-aware FusionNet is shown in Fig.~\ref{fig:fusion}. FusionNet achieves shadow free recovery by `smartly' selecting position-specific over-exposure pixels. The input data includes brackets of multiple exposure images, \ie, the shadow image $\mathbf{I}^\text{s}$, corresponding shadow mask $\mathbf{I}^\text{m}$, and over-exposure images $\{\mathbf{I}_i^\text{o}\}$. FusionNet generates fusion weight maps, across all over-exposure images, to `smartly' fuse the proper pixels from over-exposure versions with the shadow ones to the shadow free counterpart. Shadow mask $\mathbf{I}^\text{m}$ acts as a fusion guidance for FusionNet to let it assign low weights to non-shadow region and focus mostly on the shadow region, which is shown by the fusion weight maps in Fig.~\ref{fig:fusion}. 

We employ $L_1$ distance to optimize our shadow-aware FusionNet. The loss function $\mathcal{L}_\mathrm{pix}(\hat{\mathbf{I}}, \hat{\mathbf{I}}^*)$ is the pixel-wise $L_1$ distance between the ground truth shadow free image $\hat{\mathbf{I}}^*$ and the shadow removed image $\hat{\mathbf{I}}$
%
%---------------------------
\begin{align}\label{eq:l1}
\mathcal{L}_\mathrm{pix}(\hat{\mathbf{I}}, \hat{\mathbf{I}}^*) =\|\hat{\mathbf{I}}^*-\hat{\mathbf{I}}\|_{1}.
\end{align}
%
%---------------------------
%

%---------------------------------------------------------------
%---------------------------------------------------------------
\subsection{Boundary-aware RefineNet}\label{subsec:refine}
\begin{figure}[t]
\centering
\includegraphics[width=0.95\columnwidth]{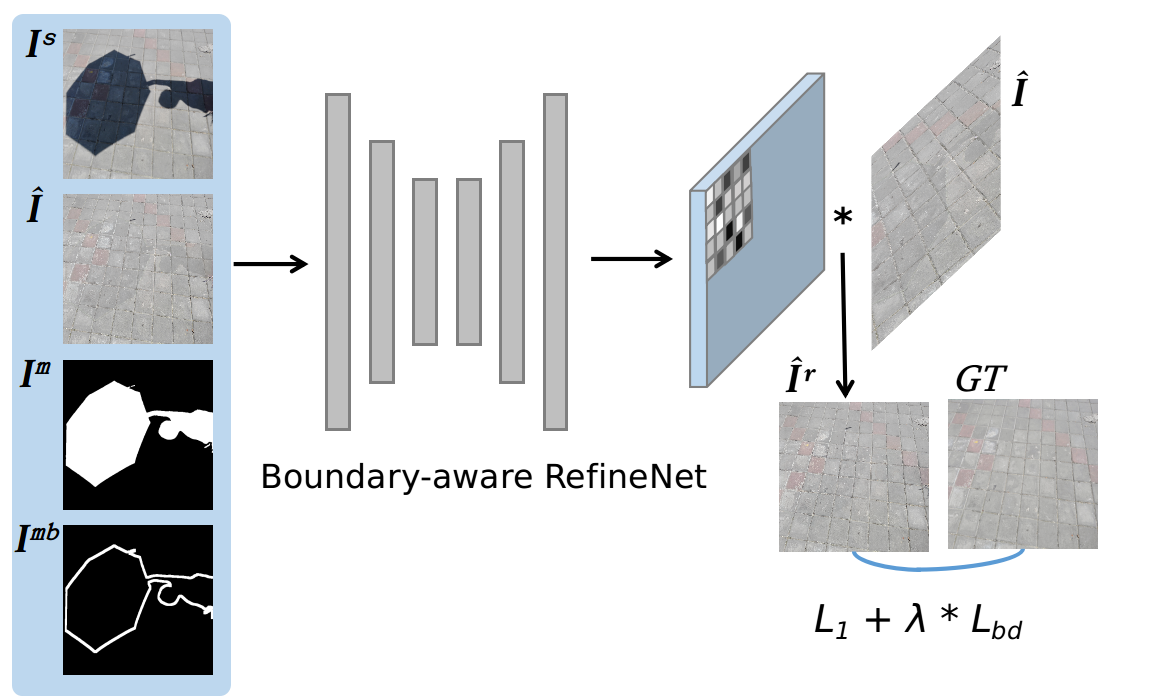}
\caption{Illustration of the proposed boundary-aware RefineNet.}
\label{fig:refine}\vspace{-12pt}
\end{figure}

Partially shadowed (penumbra) pixels exist along the shadow boundary. Inconsistent shadow patterns along the shadow boundary and inside the shadow region are still a challenge to state-of-the-art solutions to obtain traceless background. To solve this issue, we propose a boundary-aware RefineNet to eliminate the remaining shadow trace, which is shown in Fig.~\ref{fig:refine}. It acts as a refinement of the shadow removal result obtained from FusionNet. Specifically, we model the boundary-aware RefineNet as 
%
%--------------------------------
\begin{align}\label{eq:refinenet}
\mathcal{F}= \text{RefineNet}(\mathbf{I}^\text{s}, \mathbf{I}^\text{m},  \mathbf{I}^\text{mb}, \hat{\mathbf{I}}),  % change the symbol? 
\end{align}
%--------------------------------
%
where $\mathbf{I}^\text{mb}$ is a penumbra mask, as shown in Fig.~\ref{fig:refine}. Similar to Eq.~\eqref{eq:fusionv2}, $\mathcal{F}$ is also pixel-wise refine kernels that integrate the context of pixel's $k \times k$ neighborhood region with that pixel to remove remaining trace. Then the refined shadow free image becomes
%
%--------------------------------
\begin{align}\label{eq:refine}
\hat{\mathbf{I}}^\text{r}[p]= (\mathbf{F}\circledast\hat{\mathbf{I}})[p]=\sum_{q\in\mathcal{N}(p)}\mathbf{f}^{p}[p-q]\hat{\mathbf{I}}[q]
\end{align}
% &~\text{subject~to}~\sum_{i=0}^{N}\sum_{q\in\mathcal{N}(p)}\mathbf{k}_i^{p}[p-q]=1, \nonumber
%--------------------------------
%
where $\mathbf{f}^{p}\in\mathbb{R}^{k\times k}$ is the exclusive kernel for performing convolution between the $k\times k$ neighboring pixels of the pixel $p$ (\ie, $\mathcal{N}(p)$) and the kernel weights in $\mathbf{f}^{p}$.

% refine net figure
%

RefineNet's input data includes: the shadow image $\mathbf{I}^\text{s}$, shadow mask $\mathbf{I}^\text{m}$, penumbra mask $\mathbf{I}^\text{mb}$, and initial shadow removal result $\hat{\mathbf{I}}$. Penumbra mask acts as a guidance for RefineNet to keep color and illumination consistency of the shadow removed, shadow boundary, and the non-shadow regions. Penumbra mask $\mathbf{I}^\text{mb}$ is extracted by computing the difference between dilated shadow mask $\mathbf{I}^\text{md}$ and eroded shadow mask $\mathbf{I}^\text{me}$ for the penumbra region. We dilate/erode the shadow mask by 7 pixels to generate $\mathbf{I}^\text{md}$ and $\mathbf{I}^\text{me}$. The goal of RefineNet is to output a refined shadow removal image without trace.
The pixel-wise $L_1$ distance $\mathcal{L}_\mathrm{pix}(\hat{\mathbf{I}}^\text{r}, \hat{\mathbf{I}}^*)$, between the ground-truth shadow-free image $\hat{\mathbf{I}}^*$ and the refined version of shadow removed image $\hat{\mathbf{I}}^\text{r}$, is utilized to optimize the boundary-aware RefineNet. In addition, inspired by Poisson image editing~\cite{perez2003poisson}, we propose a boundary-aware loss $\mathcal{L}_\mathrm{bd}(\hat{\mathbf{I}}^\text{r}, \hat{\mathbf{I}}^\text{s}, \hat{\mathbf{I}}^*, \mathbf{I}^\text{m})$ to seamlessly remove the shadow. It is defined as
%
%---------------------------
% 要说明这里的I^m是dilate后的吗？？ 
% lan: dilate 了几个像素？7
\begin{equation}
\begin{split}\label{eq:lbd}
\mathcal{L}_\mathrm{bd}(\hat{\mathbf{I}}^\text{r}, \hat{\mathbf{I}}^\text{s}, \hat{\mathbf{I}}^*, \mathbf{I}^\text{m}) &= \mathrm{MSE}(\nabla \hat{\mathbf{I}}^\text{r}, \nabla \hat{\mathbf{I}}^\text{s})* ( 1 - \mathbf{I}^\text{m}) \\
&+  \mathrm{MSE}(\nabla \hat{\mathbf{I}}^\text{r}, \nabla \hat{\mathbf{I}}^*)* \mathbf{I}^\text{m}
\end{split}
\end{equation}
%---------------------------
%
where $\nabla$ denotes the Laplacian gradient operator. It aims to minimize the gradient domain along the shadow boundary.
It keeps the same gradient domain of non-shadow region between predicted shadow-free image $\hat{\mathbf{I}}^\text{r}$ and shadow image $\hat{\mathbf{I}}^\text{s}$. At the same time, it reduces the difference of gradient domain between predicted shadow-free image $\hat{\mathbf{I}}^\text{r}$ and ground-truth one $\hat{\mathbf{I}}^*$ in the shadow region. The total loss of RefineNet is a weighted sum of $\mathcal{L}_\mathrm{pix}(\hat{\mathbf{I}}^\text{r}, \hat{\mathbf{I}}^*)$ and $\mathcal{L}_\mathrm{bd}(\hat{\mathbf{I}}^\text{r}, \hat{\mathbf{I}}^\text{s}, \hat{\mathbf{I}}^*, \mathbf{I}^\text{m})$, as shown in Fig.~\ref{fig:refine}. We set $\lambda$ to 0.1 in the experiment.

%---------------------------------------------------------------
%---------------------------------------------------------------
\subsection{Implementation Details} \label{subsec:impl}
% Our proposed auto-exposure fusion pipeline is implemented in PyTorch. Exposure estimation is used to generate multi-exposure images as training data for shadow-aware FusionNet. Shadow-aware FusionNet is to fuse the shadow image and multi-exposure images with pixel-wise fusion weights. Then an initial shadow removed image is acquired, which will be fed into the boundary-aware RefineNet for remaining boundary trace removal to obtain a desired shadow free image. For the network setting and training, we explain them as following:

The proposed pipeline is implemented in PyTorch. The details of network setting and training are:

% \noindent
1) Exposure estimation is trained together with FusionNet. Its goal is to estimate the median-exposure version of the input shadow image. We employ ResNeXt \cite{xie2017aggregated} as backbone to do the estimation. We set the number of over-exposure images $N$ to 5 by linearly interpolating the estimated exposure parameters with scaling factors between [0.95, 1.05]. For FusionNet, we employ a DNN with U-Net256 \cite{ronneberger2015u} as backbone. 
% We train these two networks together to initially remove shadow by multi-exposure images fusion.

% \noindent
2)  Then boundary-aware RefineNet is to improve the shadow removal result with the same backbone as FusionNet. We train the RefineNet with exposure estimation and FusionNet together but freezing the latter two. Both FusionNet and RefineNet take the shadow mask as input, and we describe the setting of the datasets in Sec.~\ref{subsec:dataset}.

In our experiments, same training parameters setting are employed for these three parts. The input image is resized to 256$\times$256. The minibatch size is 8 and the initial learning rate is set to 0.0001. We use Adam optimizer for all the networks. We trained 400 epochs for each network.

%%%%%%%%%%%%%%changqing please help to fill out this

%-----------------result on ISTD and comparison results--------------------
%----------------------------------
\begin{table}
% \small
\footnotesize
\caption{Shadow removal results of our networks compared to state-of-the-art shadow removal methods on the ISTD \cite{wang2018stacked} dataset.} \label{tab:istd_data}
\centering
\footnotesize
% \resizebox{1\linewidth}{!}{
\begin{tabular}{l|cccc}
\toprule
Method ~~$\backslash$~~ \text{RMSE} & \textbf{Shadow} & \textbf{Non-Shadow} & \textbf{All}  \\
\midrule
Input Image                          & 32.12 & 7.19 & 10.97  \\
\midrule
Guo \etal ~\cite{guo2012paired}      & 18.95 & 7.46 & 9.30 \\
Gong \etal ~\cite{zhang2015shadow}  & 14.98  & 7.29 & 8.53 \\
MaskShadow-GAN ~\cite{hu2019mask}    & 12.67 & 6.68 & 7.41 \\
ST-CGAN~\cite{wang2018stacked}       & 10.33 & 6.93 & 7.47 \\
DSC~\cite{hu2019direction}           & 9.76  & 6.14 & 6.67 \\
DHAN~\cite{cun2020towards}         & 8.14  & 6.04 & 6.37 \\
% ARGAN~\cite{ding2019argan}           & 7.21$^*$ & 5.83$^*$ &  6.68$^*$ \\
\midrule
\textbf{Ours}                        & \textbf{7.77} & \textbf{5.56} &  \textbf{5.92}\\
\bottomrule
\end{tabular} 
% }
\vspace{-12pt}
\end{table}
%----------------------------------

%----------------------------------------------------------------------
%----------------------------------------------------------------------
\section{Experiments}

\subsection{Datasets and evaluation measurement}\label{subsec:dataset}

\textbf{Datasets.} We train and evaluate the proposed method on three public datasets: ISTD \cite{wang2018stacked}, adjusted ISTD (ISTD+) \cite{le2019shadow}, and SRD \cite{qu2017deshadownet} datasets. They all have paired shadow and shadow-free images. Dataset ISTD and its adjusted version also have shadow masks. We introduce these three datasets as following:

1) The training set of ISTD dataset has 1,330 triplets of shadow, shadow free, and shadow mask images. The testing split consists of 540 triplets. The ISTD+ dataset has the same number of triplets with ISTD except that it adjusts the color inconsistency, between the shadow and shadow free image, with image processing algorithm \cite{le2019shadow}. The color mismatch results from the data acquisition setup. We use ground-truth shadow masks for training stage, while for inference, we compute the shadow masks by operating Otsu's algorithm to the difference between shadow and shadow free images, similar to MaskShadow-GAN \cite{hu2019mask}. We additionally refine these masks by a median filter to remove noises.

2) SRD dataset consists of 408 pairs of shadow and shadow free images without the ground-truth shadow mask. 
%Here we use the public SRD shadow masks provided by DHAN \cite{cun2020towards} for training and testing our proposed method.
Here we simply use an adaptive threshold detection method, same as the one used in ISTD dataset, to extract the shadow mask from the difference between shadow free and shadow images. The extracted shadow masks are used both for training and testing. We utilize the public shadow masks provided by DHAN \cite{cun2020towards} for evaluation.

\textbf{Evaluation measures.} We utilize the root mean square error (RMSE) in LAB color space between the shadow removal result and the ground-truth shadow free image to evaluate the shadow removal performance, following previous works \cite{wang2018stacked, guo2012paired, qu2017deshadownet, le2019shadow, cun2020towards}. We directly compare our auto-exposure fusion framework with several state-of-the-art methods on the ISTD, ISTD+, and SRD datasets in quantitative and qualitative ways.

%----------------------------------------------------------------------
%----------------------------------------------------------------------
\subsection{Shadow removal evaluation on ISTD dataset}

We first report the shadow removal results of our method on ISTD dataset \cite{wang2018stacked}, as shown in Table~\ref{tab:istd_data}.  
We compare the proposed method with the state-of-the-art algorithms: Guo \etal~\cite{guo2012paired}, Gong \etal \cite{gong2016interactive}, ST-CGAN \cite{wang2018stacked}, MaskShadow-GAN~\cite{hu2019mask}, DSC \cite{hu2019direction}, and DHAN \cite{cun2020towards}. Different from other methods, MaskShadow-GAN utilizes unpaired shadow and shadow free images for training. The first row shows the RMSE values of the input shadow image and corresponding shadow free image without shadow removal operation. It shows that the proposed method obtains the best shadow removal performance in both shadow and non-shadow regions, leading to the lowest RMSE in the whole image. Specifically, the proposed method outperforms DSC \cite{hu2019direction} by 20.3\% and 11.2\% RMSE decreasing in shadow region and the whole image, respectively. The proposed method also outperforms the method DHAN \cite{cun2020towards} by reducing the RMSE from 8.14 to 7.77 in the shadow region.
%DHAN \cite{cun2020towards} achieved higher RMSE in the shadow region with 8.14, while the proposed method decrease it to 7.77. 
Training with unpaired data doesn't perform as well as training with paired version. Specifically, the proposed method outperforms MaskShadow-GAN by 38.6\% and 20.1\% RMSE decreasing in the shadow region and the whole area, respectively.
%

%----------------------------------
%-----------------result on ISTD+ and comparison results--------------------
%----------------------------------
\begin{table}
\footnotesize
\caption{Shadow removal results of our networks compared to state-of-the-art shadow removal methods on the ISTD+ \cite{le2019shadow} dataset.} \label{tab:istd+_data}
\centering
\footnotesize
% \resizebox{1\linewidth}{!}{
\begin{tabular}{l|cccc}
\toprule
Method ~~$\backslash$~~ \text{RMSE}  &  \textbf{Shadow} & \textbf{Non-Shadow} & \textbf{All}  \\
\midrule
Input Image & 40.2 & 2.6 & 8.5  \\
\midrule
% Yang \etal \cite{yang2012shadow}     & 24.7 & 14.4 & 16.0 \\   
Guo \etal \cite{guo2012paired}       &  22.0 & 3.1 & 6.1 \\
Gong \etal \cite{gong2016interactive}& 13.3 & - & - \\
ST-CGAN \cite{wang2018stacked}       & 13.4 & 7.7 & 8.7 \\
DeshadowNet \cite{qu2017deshadownet} & 15.9 & 6.0 & 7.6 \\
MaskShadow-GAN \cite{hu2019mask}     & 12.4 & 4.0 & 5.3 \\
Param+M+D-Net \cite{le2020shadow}    & 9.7 & \textbf{3.0} & 4.0 \\
SP+M-Net \cite{le2019shadow}         & 7.9 & 3.1 & \textbf{3.9} \\
\midrule
\textbf{Ours}                        & \textbf{6.5} & 3.8 & 4.2 \\
\bottomrule
\end{tabular} 
% }
\vspace{-12pt}
\end{table}
%----------------------------------
%
We also report the shadow removal performance of our proposed method on the adjusted ISTD (ISTD+) \cite{le2019shadow} dataset. As shown in Table~\ref{tab:istd+_data}, we compare the proposed method with state-of-the-art algorithms: Guo \etal~\cite{guo2012paired}, Gong \etal \cite{gong2016interactive}, ST-CGAN \cite{wang2018stacked}, DeshadowNet \cite{qu2017deshadownet}, MaskShadow-GAN \cite{hu2019mask}, Param+M+D-Net \cite{le2020shadow}, and SP+M-Net \cite{le2019shadow}. 
It turns out that the proposed method achieves the best shadow removal performance in the shadow region, outperforming SP+M-Net by 17.7\% lower RMSE. It outperforms the DeshadowNet and ST-CGAN trained with paired shadow and shadow-free images, decreasing the RMSE by 59.1\% and 51.4\%, respectively. Compared to methods training with unpaired data, training with paired images still acquire better results. 
The proposed method outperforms Param+M+D-Net by about 32.9\%, trained with unpaired shadow and shadow free patches. The proposed method achieves the comparable performance in the non-shadow and whole image region.
% The performance of the non-shadow region does not perform better than SP+M-Net and Param+M+D-Net, because the proposed method tends to utilize shadow mask as a guided filter for image enhancement. Multiple over-exposure could enhance the image quality for non-shadow region of shadow image, as shown in Fig.. The higher RMSE in non-shadow region does not mean bad image quality of the predicted shadow free image.
%

%
% fig for the final visualization with comparison methods
% add the result of DHAN
%
\begin{figure*}
\centering
\includegraphics[width=0.9\linewidth]{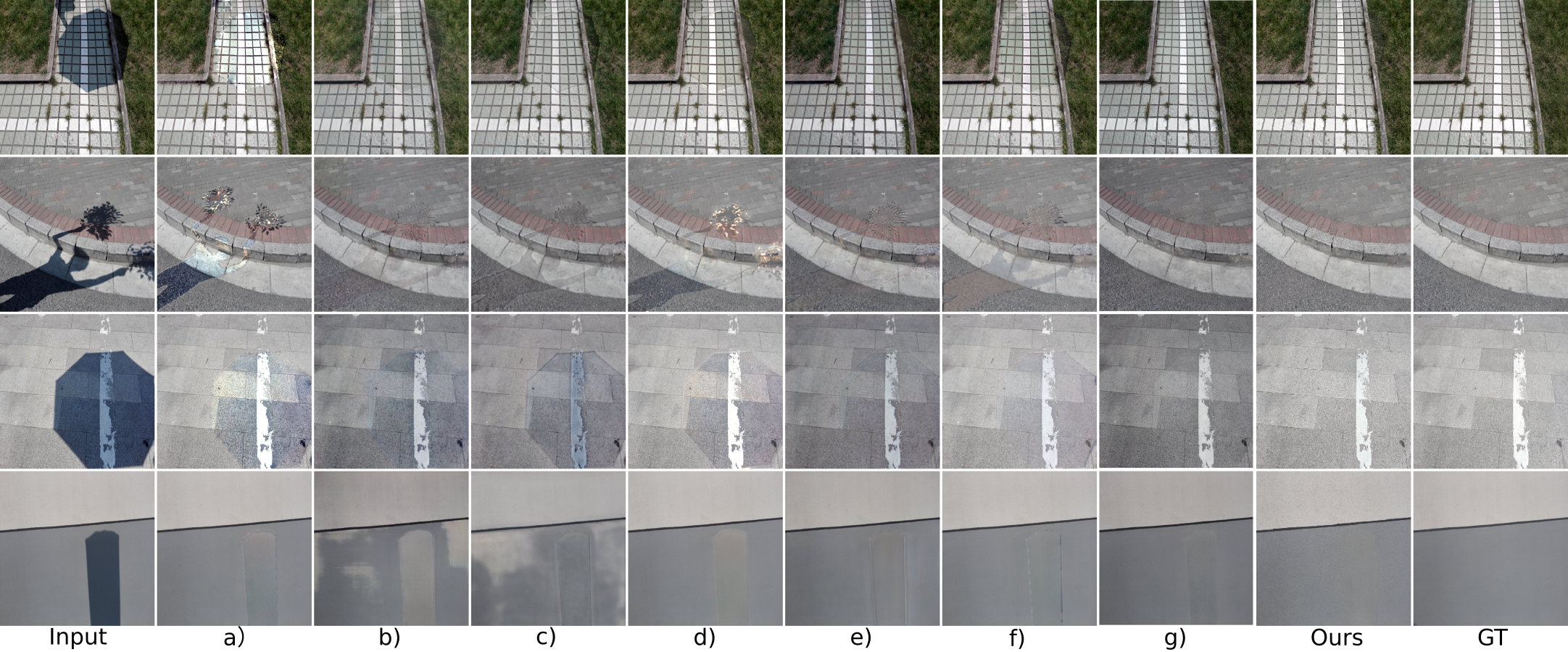}
\caption{Illustration of the visualization results of shadow removal on dataset ISTD \cite{wang2018stacked}. a) to g) are the results from comparison methods: Guo \etal \cite{guo2012paired}, ST-CGAN \cite{wang2018stacked}, MaskShadow-GAN \cite{hu2019mask}, Param+M+D-Net \cite{le2020shadow}, DSC \cite{hu2019direction}, SP+M-Net \cite{le2019shadow}, and DHAN \cite{cun2020towards}, respectively.} 
\label{fig:vis_compare}
% \vspace{-5pt}
\end{figure*}

%-----------------result on SRD and comparison results--------------------
%----------------------------------
\begin{table}[t]
\small
\caption{Shadow removal results of our networks compared to state-of-the-art shadow removal methods on the SRD \cite{qu2017deshadownet} dataset.} \label{tab:srd}
\centering
\footnotesize
% \resizebox{1.0\linewidth}{!}{
\begin{tabular}{l|cccc}
\toprule
Method ~~$\backslash$~~ \text{RMSE} & \textbf{Shadow} & \textbf{Non-Shadow} & \textbf{All}  \\
\midrule
Input Image & 40.28 & 4.76 &  14.11 \\
\midrule
Guo \etal \cite{guo2012paired}      & 29.89 & 6.47 & 12.60 \\
DeshadowNet \cite{qu2017deshadownet}& 11.78 & 4.84 & 6.64 \\
% DSC \cite{hu2019direction}          & - & - & \textbf{6.21} \\
% \midrule
DSC \cite{hu2019direction}          & 10.89 & 4.99 & 6.23 \\
DHAN~\cite{cun2020towards}        & 8.94 & \textbf{4.80} & \textbf{5.67} \\
% \midrule
% DSC \cite{ding2019argan}          & 11.31$^*$ & 6.72$^*$ &  7.83$^*$ \\
% ARGAN \cite{ding2019argan}          & 7.24$^*$ & 4.71$^*$ &  5.74$^*$ \\
\midrule
\textbf{Ours}                       & \textbf{8.56} & 5.75  & 6.51 \\ % need add experiments
%\textbf{Ours}                       & \textbf{8.01} & 6.08  & 6.51 \\ % need add experiments
\bottomrule
\end{tabular} 
\vspace{-12pt}
% }
\end{table}
%----------------------------------

%
Figure~\ref{fig:vis_compare} shows the visualization results of shadow removal from our methods and other state-of-the-art methods on the ISTD dataset. We can see that our result could recover traceless background in the shadow region. We can clearly see that traditional method, Guo \etal \cite{guo2012paired}, suffers from severe artifacts and could not recover shadowed pixels successfully due to limited feature representation ability. ST-CGAN could improve the performance by training large-scale data, while it tends to generate blurry images, random artifacts, and incorrect colors, \eg, the fourth row shadow removed image. MaskShadowGAN and Param+M+D-Net also suffer from producing blurry images. Random artifacts along the shadow boundary can be easily spotted in the result of Param+M+D-Net, and it relights the boundary rather than removing it. Even though DSC and SP+M-Net could remove most of the shadow, their results still have trace along the shadow boundary, which does not exist in our result.
%

%---------------------------------------------------------------
\subsection{Shadow removal evaluation on SRD dataset}
In this section, we show our shadow removal results on SRD dataset \cite{qu2017deshadownet} in Table~\ref{tab:srd}. We evaluate our result with the public masks provided by DHAN \cite{cun2020towards}. The proposed method obtains the best shadow removal results with the lowest RMSE in the shadow region. It reduces the RMSE from 8.94 to 8.56, compared to DHAN. 

As shown in Table~\ref{tab:srd}, the non-shadow region's RMSE values of different methods are very close (mean: 5.4, standard deviation: 0.6), which are similar to those of the Table~\ref{tab:istd+_data} for ISTD+ dataset (mean: 4.4, standard deviation: 1.7). However, the standard deviations of the RMSE values in shadow region are significantly larger. This means that different methods including ours all perform well and very close on the non-shadow region, and the main difficulty of this problem comes from the shadow region. For the shadow region, our method obviously obtains the best performance.   

% Table2 nonshadow mean: 4.385714285714286 std:1.6608333290666994
% Table3 nonshadow mean: 5.369999999999999 std: 0.6487834769782596

%%%%%%%%%%%%%%%%%%%need to update according to the new evaluate

%-----------------------------------------------------------
%-----------------------------------------------------------
\subsection{Ablation study}\label{subsec:abl}
We conduct ablation studies on ISTD+ dataset to evaluate the contribution of each step of our proposed method. 
For the effectiveness of per-pixel kernel fusion, \ie, Eq.~\eqref{eq:fusionv2} over Eq.~\eqref{eq:fusion}, we perform $\textit{Fusion-N1}$ which fuses pairs of over-exposure and shadow images with the per-pixel kernel that considers $3\times 3$ neighboring pixels and with pixel-wise fusion. It turns out that fusing image pair with neighboring information can boost the performance from 7.6 to 7.1 for RMSE in the shadow region, because neighboring region
provides important spatial context information to represent the structure. We set $3\times3$ neighborhood for FusionNet.

Then, we conduct experiments to verify the effectiveness of multiple over-exposure by controlling the number of over-exposure images. In our implementation, we set the number $N$ to 1, 3, and 5. The shadow removal models are denoted as $\textit{Fusion-N1}$, $\textit{Fusion-N3}$, and $\textit{Fusion-N5}$, respectively. $N$ is set to 5 for the remaining experiments. We test the effectiveness of boundary-aware RefineNet and loss $\mathcal{L}_\mathrm{bd}$ by models $\textit{Fusion+RefineNet}$ and $\textit{Fusion+RefineNet+$\mathcal{L}_\mathrm{bd}$}$, respectively. The results are summarized in Table~\ref{tab:abl}.

%-----------------result of ablation study--------------------
%----------------------------------
\begin{table}
\footnotesize
\caption{Ablation study of shadow removal on the ISTD+ \cite{le2019shadow} dataset.} \label{tab:abl}
\centering
\footnotesize
% \resizebox{1\linewidth}{!}{
\begin{tabular}{l|cccc}
\toprule
Method ~~$\backslash$~~ \text{RMSE} &  \textbf{Shadow} & \textbf{Non-Shadow} & \textbf{All}  \\
\midrule
Input Image & 40.2 & 2.6 & 8.5  \\
\midrule
\textit{Fusion-N1} &7.1  &3.9 & 4.4 \\   
\textit{Fusion-N3} &7.2  & 3.9 & 4.5 \\
\textit{Fusion-N5} & 6.9 & 4.0 & 4.4\\
\midrule
\textit{Fusion+RefineNet}     & 6.6 & 3.8 & 4.3\\
\textit{Fusion+RefineNet+$\mathcal{L}_\mathrm{bd}$} & \textbf{6.5}  & \textbf{3.8} & \textbf{4.2} \\
\bottomrule
\end{tabular}
% }
\vspace{-12pt}
\end{table}
%----------------------------------

%
To estimate the effectiveness of multiple over-exposure to the shadow-aware FusionNet, we report the performance in shadow, non-shadow, and whole image regions with the metric RMSE. 
When $N$ is 5, the shadow removal result in the shadow region reaches lower RMSE 6.9, compared to when $N$ = 1. We set $N$ to 5 for later ablation experiments. With the introducing of boundary-aware RefineNet, $\textit{Fusion+RefineNet}$ improves the shadow removal performance by about 0.3 RMSE decreasing. It verifies that penumbra region is a challenge for shadow removal task to get traceless background. 
% We show some examples of our result benefiting from boundary-aware refineNet in Fig.~\ref{fig:refine_vis}.
The RMSE in non-shadow region also decreased, compared to $\textit{Fusion-N5}$. Further, $\mathcal{L}_\mathrm{bd}$ loss optimized the shadow removal model $\textit{Fusion+RefineNet+$\mathcal{L}_\mathrm{bd}$}$ better to reach the lowest RMSE 6.5, 3.8, and 4.2 in the shadow, non-shadow and the whole image regions.

%----------------------------------
% refine vis
% 
% \begin{figure}
% \centering
% \includegraphics[width=\columnwidth]{image/refine_vis.png}
% \caption{Visualization results of b) $\textit{Fusion-N5}$ and c) $\textit{Fusion+RefineNet}$, respectively. a) is the corresponding shadow image.}
% \label{fig:refine_vis}
% % \vspace{-10pt}
% \end{figure}

%
%------------------------------
\begin{figure}[b]
\vspace{-10pt}
\centering
\includegraphics[width=\columnwidth]{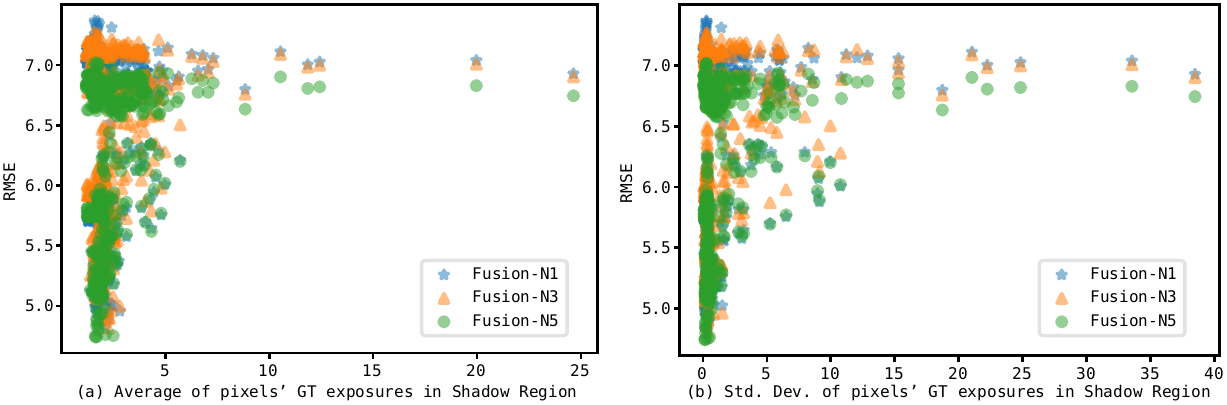}
\caption{RMSE vs. average (\ie, (a)) and std. dev. (\ie, (b)) of pixels' GT exposures in shadow region for each testing example.}
\label{fig:gtexp}\vspace{-12pt}
\end{figure}
%------------------------------

%
To explain the small margin of shadow removal performance gain of $\textit{Fusion-N5}$ over $\textit{Fusion-N1}$, we calculate the ground truth exposure for the $p$-th pixel in the shadow region by dividing the shadow-free pixel with its shadow counterpart for each testing example. 
% (\ie,  $\alpha_\text{gt}[p]=\mathbf{I}^{\text{s}}[p]/\hat{\mathbf{I}}^*[p]$)
%
Then, we count the average and the standard deviation (std. dev.) of GT exposures of all pixels in the shadow region for each example and show their relationship to the example's RMSE of the 3 variants in Fig.~\ref{fig:gtexp}.
We see that: 1) For the most examples, $\textit{Fusion-N5}$ and $\textit{Fusion-N3}$ have lower RMSE than $\textit{Fusion-N1\&N3}$ and  $\textit{Fusion-N1}$, respectively. 2) Most examples' GT exposures have small variations (\ie, small std. dev.) across spatial coordinates, leading to similar RMSE on the three methods. 3) When the GT exposures' variation become larger, the advantages of $\textit{Fusion-N3\&5}$ become more obvious.
%

%
%-----------------result of penumbra study--------------------
%----------------------------------
\begin{table}
\vspace{8pt}
\footnotesize
\caption{Comparison of traceless background results in penumbra region on ISTD+ \cite{le2019shadow} dataset.} \label{tab:penu}
\centering
\footnotesize
% \resizebox{0.6\linewidth}{!}{
\begin{tabular}{l|cc}
\toprule
Method ~~$\backslash$~~ \text{RMSE} & \textbf{Penumbra} \\
\midrule
SP+M-Net \cite{le2019shadow} & 7.06  \\   
Ours & \textbf{5.96}  \\
\bottomrule
\end{tabular}
% }
\vspace{-12pt}
\end{table}
%------------------------------

%
We also compare our method with the state-of-the-art method SP+M-Net \cite{le2019shadow} about measuring the shadow removal result without residual trace. We evaluate the RMSE metric in the penumbra region by considering the penumbra mask $\mathbf{I}^\text{mb}$ as mentioned in Sec.~\ref{subsec:refine}. As shown in Table~\ref{tab:penu}, our method performs better, decreasing RMSE by 15.5\%. Visualizations are shown in Fig.~\ref{fig:vis_compare}(f) and ours. The SP+M-Net does not perform well to remove the residual trace.
%

%
% Edge comparison
% 
% \begin{figure}\vspace{-5pt}
% \centering
% \includegraphics[width=1.0\columnwidth]{image/bd_compare.png}
% \caption{Visualization results of penumbra removal of ours and SP+M-Net\cite{le2019shadow} on ISTD+ \cite{le2019shadow} set. a) is the shadow image, b) and c) are the results of SP+M-Net and the proposed method, respectively.}
% \label{fig:edge_com}
% \vspace{-8pt}
% \end{figure}

%---------------------------------------------------------------------
%---------------------------------------------------------------------
\section{Conclusion}\label{sec:concl}
In this paper, we have proposed a novel and robust over-exposure fusion method for performing shadow removal task. Multiple over-exposure, relighting each pixel with different exposures, could compensate each pixel individually to tackle position specified color and illumination degradation. It benefits the shadow removal task by recovering the natural image from the spatial variant color and illumination degradation. 
Shadow-aware FusionNet smartly fuses brackets of over-exposure shadow images with shadow image by an adaptive per-pixel kernel weight map. It helps to fully recover the background content preserving the color and illumination details. The proposed boundary-aware RefineNet further eliminates the remaining trace caused by the penumbra area along the shadow boundary. With the boundary loss added, by optimizing to preserve the non-shadow region and recover the ground-truth shadow-free area of the shadow image, our work can obtain traceless background with the state-of-the-art shadow removal performance on the ISTD, ISTD+, and SRD datasets. In future, we plan to solve the challenging video shadow removal problem.

%---------------------------------------------------------------------
%---------------------------------------------------------------------
\noindent\textbf{Acknowledgments}: 
This work was supported by the NSFC under Grant U1803264, 61672376, 62072334, and 61671325. It was also supported in part by the National Research Foundation, Singapore under its AI Singapore Programme (AISG Award No: AISG2-RP-2020-019), Singapore National Cybersecurity R\&D Program No. NRF2018NCR-NCR005-0001, National Satellite of Excellence in Trustworthy Software System No.NRF2018NCR-NSOE003-0001, and NRF Investigatorship No. NRFI06-2020-0022. We gratefully acknowledge the support of NVIDIA AI Tech Center (NVAITC) and AWS Cloud Credits for Research Award.
%
% \cite{eccv20_spark}

%---------------------------------------------------------------------
%---------------------------------------------------------------------

% \clearpage
% \newpage

{\small
%\tiny
%\footnotesize
\bibliographystyle{ieee_fullname}
\bibliography{ref}
}

\end{document}